\pdfoutput=1

\documentclass[11pt]{article}

\usepackage[preprint]{acl}

\usepackage{times}
\usepackage{latexsym}
\usepackage{amsfonts,amsmath,amssymb, bm}
\usepackage{cleveref}

\usepackage[T1]{fontenc}

\usepackage[utf8]{inputenc}

\usepackage{microtype}

\usepackage{inconsolata}

\usepackage{graphicx}

\newcommand{\cD}{\mathcal{D}}

\newcommand{\E}{\mathop{\mathbb{E}}}
\newcommand{\Dp}{D^{\text{priv}}}
\newcommand{\Dpu}{D^{\text{pub}}}

\usepackage{color-edits}
\addauthor{mb}{green}
\addauthor{ar}{blue}

\usepackage{multirow}
\usepackage{booktabs}

\title{Order of Magnitude Speedups for LLM Membership Inference}

\author{Rongting Zhang$^*$ \\
  AWS AI \\
  \texttt{rongtz@amazon.com} \\\And
  Martin Bertran$^*$ \\
  AWS AI  \\
  \texttt{maberlop@amazon.com} \\\And
  Aaron Roth  \\
  AWS AI  and University of Pennsylvania \\
  \texttt{aaronrot@amazon.com} \\}

\begin{document}
\maketitle
\def\thefootnote{*}\footnotetext{These authors contributed equally to this work}\def\thefootnote{\arabic{footnote}}
\begin{abstract}

Large Language Models (LLMs) have the promise to revolutionize computing broadly, but their complexity and extensive training data also expose significant privacy vulnerabilities. One of the simplest privacy risks associated with LLMs is their susceptibility to membership inference attacks (MIAs), wherein an adversary aims to determine whether a specific data point was part of the model’s training set. 
Although this is a known risk, state of the art methodologies for MIAs rely on training multiple computationally costly `shadow models', making risk evaluation prohibitive for large models. 
Here we adapt a recent line of work which uses quantile regression to mount membership inference attacks; we extend this work by proposing a low-cost MIA that leverages an ensemble of small quantile regression models to determine if a document belongs to the model's training set or not. We demonstrate the effectiveness of this approach on fine-tuned LLMs of varying families (OPT, Pythia, Llama) and across multiple datasets. Across all scenarios we obtain comparable or improved accuracy compared to state of the art `shadow model' approaches,  with as little as 6\% of their computation budget.  We demonstrate increased effectiveness across multi-epoch trained target models, and architecture miss-specification robustness, that is, we can mount an effective attack against a model using a different tokenizer and architecture, without requiring knowledge on the target model. 

\end{abstract}

\section{Introduction}

 Membership inference attacks (MIAs) and reconstruction attacks pose significant risks to the privacy  of data used to fine-tune large language models (LLMs). When general purpose LLMs are used in specific applications such as automated customer support, they often require fine-tuning on proprietary, domain-specific datasets to improve their performance and relevance. This process, however, can inadvertently expose sensitive information. MIAs witness this vulnerability by reliably determining whether a specific data point was part of the training dataset or not, thereby potentially revealing personal or proprietary information.

Membership inference attacks are used to audit the privacy of trained models, and  successful external attacks can lead to breaches of confidentiality, financial loss, and erosion of user trust. Fine-tuning can amplify these risks, as models trained on smaller, specialized datasets are more susceptible to memorizing and revealing specific data points, and specialized datasets not found on the open internet can contain sensitive user information. Recent studies have found that credible privacy attacks can be mounted against modern LLMs \cite{carlini2021extracting, carlini2022membership, mattern-etal-2023-membership}. However, it is relatively uncommon to routinely assess fine-tuned LLMs for MIA risk, as current state of the art MIAs require the training of several `shadow models' --- models that are, ideally, identical in nature to the model under attack in terms of architecture, training data distribution, and hyperparameters \cite{shokri2017membership, carlini2022membership, sablayrolles2019white,watson2021importance}. The result is that mounting such an attack --- the building block of an audit --- is substantially more expensive than training the LLM in the first place.

To circumvent the high computational costs of mounting a shadow-model-based attack, recent works such as   \cite{bertran2024scalable, tang2023membership} has attempted to directly reduce the cost of hypothesis-testing style membership inference attacks by replacing shadow models with a quantile regression step that directly estimates the feature-conditional quantile of a score function from data known not to have been used in model training, where the target quantile of the score distribution directly corresponds to the false positive rate of the attack. The quantile regression approach is computationally attractive because it only requires training a single model (rather than multiple shadow models), and the architecture of the quantile regression model need not be related to (and indeed can be much simpler than) the architecture of the model under attack. Prior work has only demonstrated the effectiveness of this approach in relatively simple classification settings \cite{bertran2024scalable}  and for small diffusion models \cite{tang2023membership}. In this paper we extend this line of quantile-regression based attacks to large language models. We briefly summarize our contributions:

\vspace{-0.1in}

\begin{itemize}
    \itemsep-0.15em 
    \item We propose the use of low-cost regression ensembles to launch MIAs against LLMs. Each model in our ensemble can be significantly smaller and therefore cheaper to train than the model under attack, and need not use the same tokenizer or belong to the same model family. To exemplify this, we use Pythia-160m \cite{biderman2023pythia} and OPT-125m \cite{zhang2022opt} architectures to attack Pythia, Llama \cite{touvron2023llama}, and OPT models up to 7b parameters.

    \item We investigate performance across a variety of scoring objectives and across architecture and tokenizer. Overall, we find that our results are robust to the scoring function and the chosen architecture, in contrast to shadow model based approaches which are much more sensitive to architecture choices.
        
    \item We demonstrate the effectiveness of our approach at launching MIAs against LLMs in the low false positive rate regime on the challenging single epoch training setting on AG News, WikiText, and XSum datasets \cite{zhang2015character, merity2016pointer,narayan2018don}. Our approach robustly outperforms other baselines, with as little as $6\%$ of the compute required on the larger architectures.

\end{itemize}

\begin{figure}[tb!]
\centering
\includegraphics[width=0.475\textwidth]{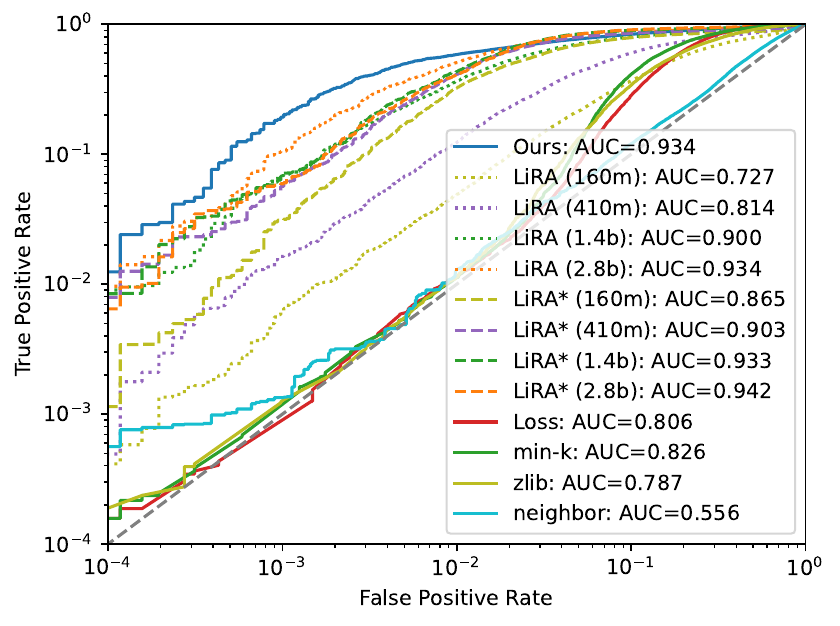}
\includegraphics[width=0.475\textwidth]{figs/pythia_6.9b_wikitext_roc-1.pdf}
\caption{Comparing true positive rates vs false positive rates of our method with LiRA variants and simple score-function-based methods on WikiText-103, where target model is Pythia-6.9b. LiRA* represents LiRA with fixed variance estimate. LiRA results are obtained with 4 shadow models from Pythia family of varying sizes. Results for our method are obtained with ensemble of 5 quantile regression models fine-tuned from Pythia-160m.}
\label{fig:pythia_6_9_b_wikitext_roc}
\vspace{-0.15in}
\end{figure}
\section{Additional Related Work}

\subsection{Membership Inference Attacks}
Membership inference aims to determine whether a sample is used in the training set of a model. 
Initiated by the seminal work ~\cite{homer2008resolving}, membership inference attacks are of central importance in  privacy risk assessment of machine learning models since successful MIA results falsify differential privacy guarantees \cite{dwork2006differential,dwork2014algorithmic} and can be substantially disclosive in their own right.
Most membership inference attacks can be viewed as hypothesis tests defined over  score functions such as training loss~\cite{yeom2018privacy}, and are based on the premise that models tend to overfit the data they were trained on and so induce a different distribution of scores on training data as compared to identically distributed data that was not used in training.

Shadow-model-based methods train multiple models similar or identical to the model under attack on datasets sampled from the same distribution as the training set, so that each shadow model can be used to produce a sample from  
the null hypothesis that a given data point was not used in training or  its corresponding alternative hypothesis;
given many trained shadow models, it is then possible to fit parametric families of distributions to the null and alternative hypothesis \cite{shokri2017membership,jayaraman2021revisiting,carlini2022membership}. \cite{carlini2022membership} in particular formalizes this hypothesis testing approach,  and proposes to use the optimal likelihood ratio test on these fit distributions.  LiRA has been shown to achieve state of the arts results in a wide variety of settings.

Because shadow models are designed to produce samples from the null or alternative hypothesis distributions, shadow-model-based methods  require knowledge of the model architecture and training process so as to be able to replicate the entire training pipeline of the model under attack. This additionally makes training each shadow model as expensive as training the model under attack.
Motivated by these limitations, \citet{bertran2024scalable} proposed a quantile regression based attack which reframes the hypothesis testing problem over the randomness of the training and test data, rather than over the randomness of the model training. \citet{bertran2024scalable} evaluate their method for classification models, and 
\citet{tang2023membership} extends this method to attack simple diffusion models. Our work is a direct extension of this line. 

\subsection{Membership Inference on LLMs}
\label{sec:scores}
Because shadow model approaches require replicating the model training process many times, they are prohibitively expensive to mount on large langauge models. As a result many recent works on membership inference attacks on LLMs focus on proposing more effective scoring functions that can be applied without calibration (i.e., using marginal thresholds). Examples include local curvature of the loss by comparing samples with neighboring texts~\cite{mattern-etal-2023-membership},
conditioned score on a subset of high perplexity tokens ~\cite{shi2024detecting}, and re-normalizing
loss by compression length under zlib~\cite{carlini2021extracting}, among others \cite{long2018understanding,watson2021importance}. While these methods produce results with minimal computational costs, their performance lags behind those that learn a calibration function.

\subsection{Memorization in LLMs}
Memorization is a related but different concern than vulnerability to membership inference, often motivated by copyright infringement. Definitions of memorization are still being actively explored. 
There are have been attempts to define memorization through prompting language models to regurgitate text with varying types of prefixes~\cite{carlini2021extracting}, ~\cite{nasr2023scalable}, ~\cite{carlini2023quantifying}, counterfactual notions of memorization ~\cite{zhang2023counterfactual}, and adversarial compression rate of text~\cite{schwarzschild2024rethinking}.

\section{Method}

Here we provide a detailed description of our attack, starting with the general idea of how score-based membership inference attacks are designed, and followed with a technical description of how we design our low-cost ensemble attack.

We follow a standard setup of membership inference attacks in which the adversary has query access to an LLM $f$ trained on an unknown dataset $\Dp$ in terms of log likelihoods per-token for an arbitrary input sequence, sampled from a document distribution $\cD$. Each sample $\bm{x} \in \Dp$ consists of a document or sentence, usually split into tokens $\bm{x} =\{x_i\}$ by a model-specific tokenizer. The model $f$ outputs a probability distribution of the next token $x_i$ conditioned on the preceding token sequence $\bm{x}_{<i} = x_1, \dots, x_{i-1}$. We let $f(x_i \mid \bm{x}_{<i})$ denote the likelihood of token $x_i$ assigned by model $f$, conditioned on the preceding tokens. A model $f$ is usually fine-tuned on a dataset $\Dp$ by minimizing negative log likelihood:

\vspace{-1em}
\begin{equation}
    \mathcal{L}(\theta, \Dp) = \sum_{\bm{x} \in \Dp } \sum_{i\in[n]} -\log f(x_i\mid \bm{x}_{<i}),
\end{equation}
\noindent with $n$ the number of tokens in $x$.

Because of this, training samples are potentially memorized, or are more likely under the model's distribution than other, similar samples that might be equally likely under the sampling distribution. 

A membership inference attack is a hypothesis test that exploits this tendency by using a test statistic (score) derived  from queries to $f$ that aims to determine whether a document $\bm{x}$ is a member of the training set $\Dp$ or not. We cast this as distinguishing between a null and alternative hypothesis: 

\vspace{-1em}
$$H_0: \bm{x} \sim \cD \ \ \ \ \ \ \ H_1: \bm{x} \sim \Dp.$$
We restrict our attention to membership inference attacks that define a test statistic (score) $s(\bm{x};f)$. These attacks determine if this input-score pair $(s(\bm{x};f), x)$ is likely under the null hypothesis, and accuse a document of being part of the private dataset if this test fails (i.e. rejects the null). We'll denote $s(\bm{x};f) = s(\bm{x}) = s$ when clear from context. A score is any function computable given access to the model $f$ and target point $\bm{x}$. The intention is to choose a score that takes systematically higher values for $\bm{x} \in \Dp$. Examples of such scores are discussed in \ref{sec:scores}; for scores that are computed \textit{per token}, we take the \textit{per document} score to be the token-averaged score.

The adversary's goal is to learn an attack function $A_f: \mathcal{X}\rightarrow \{0,1\}$ that implements the hypothesis test described above. The works discussed here follow a common thread by implementing the adversary as

\vspace{-1em}
\begin{equation}
    A_f(\bm{x}) = \mathbf{1}[s(\bm{x}) \ge q(\bm{x})],
\end{equation}
where the threshold $q(\bm{x})$ is sometimes referred to as `difficulty calibration'. Attacks are differentiated based on their choice of score function, and their choice of threshold function. For example \cite{yeom2018privacy} uses negative log likelihood as their score function, and a constant (marginal) threshold function. \cite{sablayrolles2019white,watson2021importance,carlini2022membership} use shadow models \cite{shokri2017membership} to determine a suitable per-example threshold. Notably, \cite{ carlini2022membership} proposes an `offline' test that models the score distribution of a document under $H_0$ as $\mathcal{N}(\mu(\bm{x}), \sigma(\bm{x})^2)$, where the mean and variance are the empirical mean and variances of the score function computed across all shadow models that do not include $\bm{x}$ in their training set, the threshold function $q(\bm{x})$ is then computed as a \textit{quantile} of the normalized score distribution $q(\bm{x}) = \phi^{-1}(1-\alpha) \sigma(\bm{x}) + \mu(\bm{x})$ with $\alpha$ a target false positive rate and $\phi^{-1}$ the inverse CDF of a standard distribution. There are other score functions such as min-k~\cite{shi2024detecting}, zlib entropy~\cite{carlini2022membership}, and neighborhood comparison attack~\cite{mattern-etal-2023-membership} that can be viewed as adaptive threshold methods. In this work we choose to characterize them as non-adaptive scores since this characterization enables further refinement using shadow models and quantile regression.

\subsection{Quantile Regression}

The recent work of \cite{bertran2024scalable} proposed to do away with shadow models by instead learning a quantile regression model to directly predict (a quantile of) the score function for public data by minimizing pinball loss for the target quantile. Here we instead build our parametric regression model as a pair of functions $\mu: \mathcal{X}\rightarrow \mathbf{R}$, $\sigma: \mathcal{X}\rightarrow \mathbf{R}^+$ that respectively predict the mean and standard deviation of the score distribution under the null hypothesis. Given a dataset $\Dpu \sim \cD$ and a family of regression models $r \in \mathcal{R}$ we minimize either

\vspace{-2em}
\begin{align}
         \tiny \E_{s,\bm{x} \sim \Dpu} [-\log \mathcal{N}(s; \mu(\bm{x}), \sigma^2(\bm{x}))], \label{eq:rmseuncertainty} \\
         \tiny \E_{\bm{x} \sim \Dpu} [PB_{\phi(0)}(\mu(\bm{x}), s) +\nonumber\\
         \qquad PB_{\phi(1)}(\mu(\bm{x})+\sigma(\bm{x}), s)]. \label{eq:rmseuncertainty}
\end{align}
\vspace{-1.5em}

Where in the first scenario we learn the mean and std of the distribution by minimizing negative log likelihood of a normal distribution, and on the latter we directly learn the median and $\phi(1)$ quantile of the distribution using pinball loss\footnote{$\textrm{PB}_{1-\alpha}(\hat y, y) = \max\{\alpha (\hat y-y), (1-\alpha)(y-\hat y)\}$}; the $\phi(1)$ quantile corresponds to a point falling below 1 standard deviation above the mean of a standard Gaussian, and thus is chosen as a natural target for a `robust' estimate of standard deviation of the score distribution. The second objective we propose shares some of the advantages of the parametric negative log likelihood approach (only two outputs are needed to model the score distribution), but instead relies on robust quantile estimators that can be used to derive mean and standard deviation of the distribution under the Gaussian assumption.

In both settings, the quantile threshold is computed as $q_{1-\alpha}(\bm{x}) = \phi^{-1}(1-\alpha) \sigma(\bm{x}) + \mu(\bm{x})$ with $\alpha$ a target false positive rate and $\phi^{-1}$ the inverse CDF of a standard distribution.

To decide which of these objectives provides a more suitable base model, we choose the one with the smallest pinball loss at the target false positive rate measured on public data, similarly to \cite{bertran2024scalable}. We note that this statistic can be computed without access to private data. The relative performance of these objectives varies by dataset, with both producing strong results.

The regressor need not have the same architecture as the model inducing the score function, in this work we opt to use an ensemble of weak learners to minimize compute costs as described next.

\subsection{Ensemble of Quantile Regression Models}

The work in \cite{tang2023membership} on MIAs against diffusion models used a bootstrapping approach in which multiple small quantile regression models ``voted'' on whether to accuse a point of membership in the training set. Our preliminary experiments showed that using the entire dataset per ensemble produced better results than bootstrapping, so here we instead choose to leverage the entire public data $\Dpu$ by using deep ensembles of (weak) learners as in \cite{lakshminarayanan2017simple}. We treat each model in the ensemble as a uniformly-weighted mixture model, and compute the mean and variance of the ensemble as 

\vspace{-1.5em}
\begin{align}
    \tiny \mu_*(\bm{x}) &= \frac{1}{M} \!\sum_{m\in [M]} \mu_m(\bm{x}),\\
     \sigma^2_*(\bm{x}) &= \frac{1}{M}\! \sum_{m\in [M]} \sigma^2_m (\bm{x}) \!+\!\mu^2_m(\bm{x}) \!-\!\mu^2_*(\bm{x}).
\end{align}

This methodology allows us to leverage more of the available public samples on each individual model of the ensemble, compared to a bootstrap approach, where roughly $63\%$ of the samples are used per ensemble\footnote{The bootstrap method uniformly resamples a dataset $\Dp_m$ with replacement for each model in the ensemble such that $|\Dp_m|=|\Dp|$}. This approach averages the mean and std of the models, and then computes the appropriate quantile given these averaged parameters, as opposed to the voting approach used in \cite{tang2023membership} in which each model of the ensemble votes on the membership of a document.


\section{Experiment Setup}
\subsection{Datasets}
We conducted experiments on three public datasets across different domains:
AG News~\cite{zhang2015character}, WikiText-103 ~\cite{merity2016pointer}, and XSum~\cite{narayan2018don}.
On WikiText-103, we sampled around 22.5\% of the full dataset and excluded examples that contain less than 25 characters.
On XSum, we took the original article as the text samples.
On each dataset, we split the data samples into two halves, where one half is used to fine-tune language models and is regarded as the private dataset. 
The other half is regarded as the public dataset and further split into two sets, which we name as public-train and public-test respectively. 
Public-train set was used to train quantile regression models for our method and shadow models for LiRA while the public-test set was used as a holdout set for testing.
Membership inference attacks were evaluated on the union of private and public-test splits.
Table \ref{tab:datasets} shows the split sizes and statistics on sample length for each dataset.

\begin{table}[th!]
\scriptsize
\centering
\begin{tabular}{cc@{\hspace{0.6em}}c@{\hspace{0.6em}}cc@{\hspace{0.6em}}c@{\hspace{0.6em}}c}
\toprule
\multirow{2}{*}{Dataset} & \multicolumn{3}{c}{\#Examples} & \multicolumn{3}{c}{Length} \\
\cmidrule(lr){2-4} \cmidrule(lr){5-7}
        & private & public-train & public-test & 25\% & 50\% & 75\% \\
\midrule
AG News & 51210 & 51577 & 11213 & 196 & 232 & 265 \\
WikiText & 101487 & 100846 & 25562 & 85 & 493 & 804 \\
XSum     & 91619 & 92169 & 20257 & 1040 & 1747 & 2910 \\
\bottomrule
\end{tabular}
\caption{\label{tab:datasets} Statistics on split size and document length of all evaluated datasets.}
\vspace{-0.15in}
\end{table}

\subsection{Target Language Models}
We considered three widely adopted families of LLMs as targets for membership inference, including
Pythia ~\cite{biderman2023pythia}, 
OPT ~\cite{zhang2022opt}
and Llama ~\cite{touvron2023llama}. 
We fine-tuned Pythia-160m for quantile regression against Pythia models and OPT-125m against OPT and Llama models. 

\subsection{Baselines}
To evaluate the performance of the proposed method, we compared it with different score function baselines without difficulty calibration, including
loss attack ~\cite{yeom2018privacy},
min-k\% ~\cite{shi2024detecting},
zlib entropy~\cite{carlini2021extracting} 
and neighborhood comparison attack~\cite{mattern-etal-2023-membership}.
We also conducted extensive comparison against LiRA~\cite{carlini2022membership} variants that use variable and fixed variance estimates.

\subsection{Implementation Details}
We fine-tuned target language models on the private split of each dataset for 3 epochs with Adam with a learning rate of $5\times 10^{-5}$ and batch size of 64; we used HuggingFace public checkpoints as a starting point for all models. Unless otherwise noted, we report MIA results on the first epoch of training, since this represents the most challenging scenario where each sample in the private set is only seen once by the target model. Shadow models used in LiRA experiments were trained on sampled subsets of the public-train split of each dataset with identical settings as the target language models.
Quantile regression models were trained on the public-train split of the dataset for 4 epochs with Adam with a learning rate of $2 \times 10^{-5}$ and batch size of 128. We stored snapshots of the quantile regression model at integer epochs and picked the snapshot with the best evaluation loss on a holdout set sampled from the public-train split. All experiments were conducted on a machine with 8 V100 GPUs.

\begin{table*}[th!]
\scriptsize	
\centering
\begin{tabular}
{@{\hspace{0.2em}}c@{\hspace{0.2em}}c@{\hspace{0.8em}}c@{\hspace{0.6em}}c@{\hspace{0.6em}}c@{\hspace{0.6em}}c@{\hspace{0.6em}}c@{\hspace{0.6em}}c@{\hspace{0.6em}}c@{\hspace{1.1em}}c@{\hspace{0.6em}}c@{\hspace{0.6em}}c@{\hspace{0.6em}}c@{\hspace{0.6em}}c@{\hspace{0.6em}}c@{\hspace{0.6em}}c@{\hspace{1.1em}}c@{\hspace{0.6em}}c@{\hspace{0.6em}}c@{\hspace{0.6em}}c@{\hspace{0.6em}}c@{\hspace{0.6em}}c@{\hspace{0.6em}}c@{\hspace{0.2em}}}
\toprule
\multirow{2}{*}{FPR} & \multirow{2}{*}{Model} & \multicolumn{7}{c}{AG News} & \multicolumn{7}{c}{WikiText} & \multicolumn{7}{c}{XSum} \\
\cmidrule(l{0pt}r{9pt}){3-9} \cmidrule(l{0pt}r{9pt}){10-16} \cmidrule(l{0pt}r{3pt}){17-23}
& & Loss & min-k & zlib & Ne & LiRA & LiRA* & Ours & Loss & min-k & zlib & Ne & LiRA & LiRA* & Ours & Loss & min-k & zlib & Ne & LiRA & LiRA* & Ours \\
\midrule
\multirow{5}{*}{0.1\%} & P-2 & 0.29 & 0.40 & 0.56 & 0.63 & 2.12 & 3.70 & \textbf{5.46} & 0.04 & 0.07 & 0.04 & 0.18 & 9.57 & 4.80 & \textbf{15.57} & 0.15 & 0.23 & 0.12 & 0.09 & 11.06 & 15.49 & \textbf{25.69} \\
& P-6 & 0.29 & 0.46 & 0.76 & 0.79 & 2.03 & 4.04 & \textbf{6.95} & 0.04 & 0.07 & 0.12 & 0.13 & 10.53 & 5.92 & \textbf{19.29} & 0.15 & 0.25 & 0.13 & 0.10 & 7.82 & 22.45 & \textbf{36.92} \\
& O-2 & 0.15 & 0.19 & 0.30 & 0.30 & 2.82 & 3.13 & \textbf{4.75} & 0.02 & 0.10 & 0.02 & 0.16 & 8.01 & 3.50 & \textbf{11.07} & 0.15 & 0.11 & 0.11 & 0.09 & 12.25 & 14.82 & \textbf{14.90} \\
& O-6 & 0.18 & 0.28 & 0.46 & 0.59 & 2.44 & 3.05 & \textbf{5.32} & 0.04 & 0.10 & 0.04 & 0.10 & 5.78 & 4.76 & \textbf{14.48} & 0.14 & 0.13 & 0.12 & 0.09 & 4.12 & 18.55 & \textbf{27.56} \\
& L-7 & 0.33 & 0.35 & 0.62 & 0.57 & 1.05 & 3.67 & \textbf{7.42} & 0.33 & 0.23 & 0.25 & 0.27 & 0.44 & 1.94 & \textbf{17.96} & 0.13 & 0.15 & 0.14 & 0.09 & 0.42 & 12.16 & \textbf{34.94} \\
\midrule
\multirow{5}{*}{1\%} & P-2 & 2.45 & 5.19 & 4.97 & 3.50 & 17.92 & \textbf{34.25} & 28.73 & 1.13 & 1.06 & 1.05 & 1.37 & \textbf{48.79} & 33.23 & 45.94 & 1.93 & 6.98 & 4.53 & 1.89 & 62.71 & \textbf{70.87} & 59.38 \\
& P-6 & 3.35 & 7.08 & 7.46 & 3.68 & 17.65 & 33.78 & \textbf{34.17} & 1.09 & 1.11 & 1.12 & 1.30 & 50.72 & 40.72 & \textbf{58.52} & 2.70 & 11.10 & 6.55 & 2.07 & 50.87 & \textbf{76.99} & 72.68 \\
& O-2 & 1.45 & 2.83 & 3.30 & 2.47 & 16.30 & \textbf{26.75} & 21.41 & 1.06 & 1.03 & 1.05 & 1.10 & \textbf{33.74} & 23.22 & 26.98 & 1.34 & 3.50 & 2.79 & 1.44 & 49.77 & \textbf{56.00} & 36.43 \\
& O-6 & 2.65 & 5.14 & 6.20 & 3.36 & 19.74 & 31.74 & \textbf{33.79} & 0.93 & 1.04 & 1.08 & 1.12 & 34.78 & 36.74 & \textbf{53.69} & 1.83 & 6.88 & 5.51 & 1.70 & 32.83 & \textbf{71.91} & 67.70 \\
& L-7 & 4.15 & 6.37 & 7.59 & 3.18 & 7.51 & 25.45 & \textbf{39.38} & 1.68 & 1.68 & 1.62 & 1.42 & 4.62 & 17.34 & \textbf{61.87} & 3.71 & 14.34 & 5.43 & 1.81 & 4.69 & 45.28 & \textbf{81.46} \\
\bottomrule
\end{tabular}
\caption{\label{tab:main}True positive rates (\%) at 0.1\% and 1\% false positive rate (FPR) of different membership inference methods on the three datasets. P-2, P-6, O-2, O-6, L-7 correspond to Pythia-2.8b, -6.9b, OPT-2.7b, -6.7b, and Llama-7b, respectively. LiRA* represents LiRA with fixed variance. All LiRA results are obtained with 4 shadow models. The shadow models are Pythia-2.8b models for Pythia models, and OPT-2.7b models for OPT and Llama models.}
\vspace{-0.2in}
\end{table*}

\section{Results}
Here we discuss our experimental results and main observations.
We first show an overall performance evaluation of our method along with the baselines. We then specifically discuss scalability and cross-family performance, and compare against LiRA in this scenario. We further explore how ensemble size affects the performance of our method. Finally we study different factors affecting the privacy risks of fine-tuned models including target model sizes and training epochs and how these impact different MIA methods. Additional experiments measuring robustness to score function selection are presented in Appendix \ref{sec:scoring_comparison}.

\subsection{Comparison with Baselines}
Table~\ref{tab:main} shows the performance of our proposed method and baselines on AG News, WikiText and XSum.
We compute the true positive rates at 0.1\% and 1\% false positive rates of all methods.
Due to compute limits, we trained 4 shadow models for each setting  and did not train shadow models from exactly the same pretrained model for larger models. For Pythia-6.9b and OPT-6.7b, we used Pythia-2.8b and OPT-2.7b as shadow models correspondingly. For Llama-7b, we picked OPT-2.7b as the shadow model architecture as it showed better performance compared to Pythia-2.8b with LiRA methods.
The results for our method were obtained using an ensemble of 5 models.

We observe that loss, min-k\%, zlib entropy, and neighborhood comparison attacks perform poorly, especially on the more challenging WikiText dataset where there is a great variety in topic and text length among the samples.
This highlights the importance of per-sample calibration in achieving high performance in low false positive regime.

In our experiments across all datasets, our method shows performance comparable to the two LiRA variants. 
It achieves the best performance among all methods at 0.1\% FPR across all datasets and model families. This illustrates the effectiveness and robustness of our method in the low false positive rate regime.
LiRA achieves strong performance at 1\% FPR across datasets when target models and shadow models are derived from exactly the same pretrained model. 
For Pythia-6.9b and OPT-6.7b, LiRA methods are outperformed by our method on AG News and WikiText. This is likely due to the mismatch in model sizes between shadow models and target models, necessitated by LiRA's very large computational requirements.
For Llama-7b, where shadow models are from a different model family, LiRA methods are outperformed by our method across all three datasets by a large margin. 
These results demonstrate the favorable performance of our method compared to LiRA with few shadow models especially when it is impractical to leverage shadow models that share the same architecture or size with the target model architecture due to limited compute or lack of information. The following section details the exact computation costs of each attack. 
Extended results on ROC curves are presented in Appendix \ref{sec:extended_roc_results}.

\begin{table}[h!]
\scriptsize	
\centering
\begin{tabular}{l@{\hspace{0.8em}}c@{\hspace{0.7em}}cc@{\hspace{0.7em}}cc@{\hspace{0.7em}}c}
\toprule
Dataset & \multicolumn{2}{c}{AG News} & \multicolumn{2}{c}{WikiText} & \multicolumn{2}{c}{XSum}\\
\cmidrule(l{0pt}r{4pt}){2-3} \cmidrule(l{4pt}r{4pt}){4-5} \cmidrule(l{4pt}r{4pt}){6-7}
FPR & 0.1\% & 1\% & 0.1\% & 1\% & 0.1\% & 1\%\\
\midrule
Loss & 0.29 & 3.35 & 0.04 & 1.09 & 0.15 & 2.70\\
\midrule
LiRA (160m) & 0.63 & 5.61 & 0.61 & 4.90 & 0.49 & 3.36\\
LiRA (410m) & 1.30 & 12.57 & 1.64 & 12.37 & 1.06 & 8.73\\
LiRA (1.4b) & 1.63 & 16.47 & 6.84 & 36.15 & 3.29 & 26.24\\
LiRA (2.8b) & 2.03 & 17.65 & 10.53 & 50.72 & 7.82 & 50.87\\
\midrule
LiRA* (160m) & 4.60 & 25.53 & 3.10 & 32.31 & 11.64 & 39.60\\
LiRA* (410m) & 4.16 & 32.97 & 5.66 & 41.21 & 18.89 & 63.67\\
LiRA* (1.4b) & 3.43 & 33.50 & 6.77 & 43.28 & 21.43 & 75.06\\
LiRA* (2.8b) & 4.04 & 33.78 & 5.92 & 40.72 & 22.45 & \textbf{76.99}\\
\midrule
Ours (160m) & \textbf{6.95} & \textbf{34.17} & \textbf{19.29} & \textbf{58.52} & \textbf{36.92} & 72.68\\
\bottomrule
\end{tabular}
\caption{\label{tab:scalability}True positive rates (\%) at 0.1\% and 1\% FPR on the three datasets; the target model is Pythia-6.9b. LiRA results are obtained with 4 shadow models from the same Pythia family with different sizes. LiRA* represents LiRA with fixed variance. }
\vspace{-0.15in}
\end{table}

\vspace{-0.1in}
\subsection{Scalability of our Attack}
Table~\ref{tab:scalability} shows a performance comparison between our method and LiRA at different shadow and regression model sizes when the target model is Pythia-6.9b.
For LiRA methods, performance generally improves with the size of shadow models, which is unsurprising since there is less difference between target and shadow models. 
The trend is particularly evident for LiRA with per-sample variance, while LiRA with fixed variance is more stable across different shadow model sizes. This indicates that the variance estimate in LiRA is significantly more sensitive to shadow model sizes compared to mean estimate, at least on this particular scenario.
To achieve competitive results with LiRA on challenging datasets such as WikiText, it would be best to use shadow models of similar sizes as target model.
In contrast, our method achieves high performance even when the size of the target model is significantly larger than the regression model.
Additional analysis on performance by regression model size of our method is shown in Appendix \ref{sec:extended_size_results}.

Table~\ref{tab:train_time} shows a comparison on time required to train a single model for the attacks on XSum, Pythia-160m regression models for our method and Pythia models up to 6.9b for LiRA. Our method requires only $6\%$ of the compute time required for LiRA with 4 Pythia-2.8b shadow models and $1.5\%$ of the time would be required if Pythia-6.9b shadow models were used.

\begin{table}[h]
\vspace{-0.1in}
\scriptsize
\centering
\begin{tabular}{ll}
\toprule
Method & Time (Hours) \\
\midrule
LiRA (Pythia-160m) & 0.75 \\
LiRA (Pythia-410m) & 1.73 \\
LiRA (Pythia-1.4b) & 12.94 \\
LiRA (Pythia-2.8b) & 13.31* \\
LiRA (Pythia-6.9b) & 53.73* \\
\midrule
Ours (Pythia-160m) & 0.64 \\
\bottomrule
\end{tabular}
\caption{Time required to train a single shadow model for LiRA and regression model for our method on XSum. The two largest Pythia models were trained using mixed precision (*). In our experiments, LiRA results were obtained with 4 shadow models while results for our method were obtained with 5-model ensembles.}
\label{tab:train_time}
\vspace{-0.25in}
\end{table}

\vspace{-0.1in}
\subsection{Cross Family Performance}
In Table~\ref{tab:cross_family}, we show a comparison of our method with LiRA on WikiText where the model family varies among target model and attacker models.
In the experiments with Pythia-6.9b and OPT-6.7b as target models, we observe that both our method and LiRA performs better when the target model and shadow models are from the same model family in general. 
However, the performance of our method is less influenced by the difference in model families.
In fact, our method with mismatched model families is able to outperform LiRA with matched model families in the experiments.
In the experiments with Llama-7b, we observe a dramatic degradation in the performance of LiRA methods.
In constrast, our method is able to achieve relative stable performance as measured by TPR at 1\% FPR with different choices of model families. 
Nonetheless, a significant difference in TPR at 0.1\% FPR is observed for our method, signifying the difficulty of maintaining competitive performance at lower false positive regime when target model architecture is not exactly known. 

\begin{table}[h!]
\scriptsize	
\centering
\begin{tabular}{l@{\hspace{0.8em}}c@{\hspace{0.7em}}cc@{\hspace{0.7em}}cc@{\hspace{0.7em}}c}
\toprule
Model & \multicolumn{2}{c}{Pythia-6.9b} & \multicolumn{2}{c}{OPT-6.7b} & \multicolumn{2}{c}{Llama-7b}\\
\cmidrule(l{0pt}r{4pt}){2-3} \cmidrule(l{4pt}r{4pt}){4-5} \cmidrule(l{4pt}r{4pt}){6-7}
FPR & 0.1\% & 1\% & 0.1\% & 1\% & 0.1\% & 1\%\\
\midrule
Loss Attack & 0.04 & 1.09 & 0.04 & 0.93 & 0.33 & 1.68\\
\midrule
LiRA (Pythia-2.8b) & 10.53 & 50.72 & 4.51 & 24.88 & 0.48 & 4.51\\
LiRA (OPT-2.7b) & 5.73 & 36.96 & 5.78 & 34.78 & 0.44 & 4.62\\
\midrule
LiRA* (Pythia-2.8b) & 5.92 & 40.72 & 3.98 & 20.59 & 1.88 & 12.35\\
LiRA* (OPT-2.7b) & 5.91 & 47.09 & 4.76 & 36.74 & 1.94 & 17.34\\
\midrule
Ours (Pythia-160m) & \textbf{19.29} & 58.52 & 13.03 & 45.98 & 11.28 & 56.74\\
Ours (OPT-125m) & 18.03 & \textbf{61.19} & \textbf{14.48} & \textbf{53.69} & \textbf{17.96} & \textbf{61.87}\\
\bottomrule
\end{tabular}
\caption{\label{tab:cross_family} True positive rates (\%) at 0.1\% and 1\% FPR with different target models on the WikiText dataset. LiRA results are obtained with 4 shadow models from the Pythia and OPT families. LiRA* represents LiRA with fixed variance. }
\vspace{-0.15in}
\end{table}

\subsection{Effect of Ensemble Size}

\begin{figure}[htb!]
\vspace{-0.15in}
\centering
\includegraphics[width=0.4\textwidth]{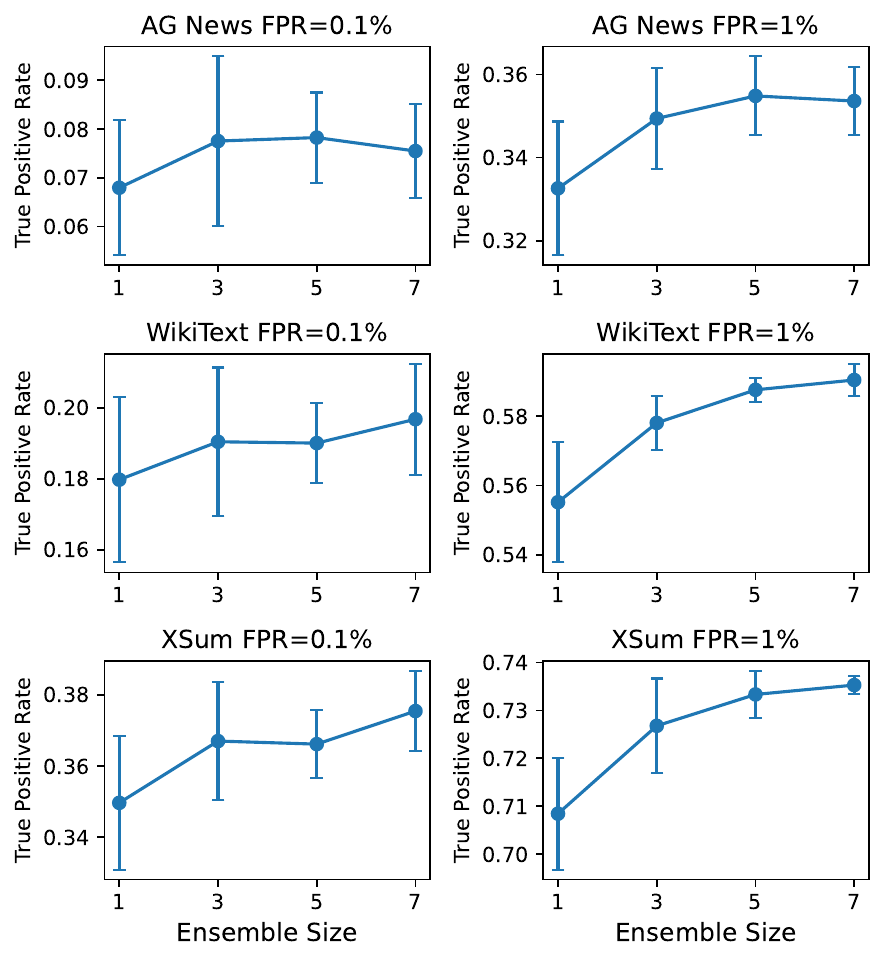}
\caption{True positive rates at 0.1\% and 1\% FPR on the three datasets where target model is Pythia-6.9b, with varying ensemble sizes of our method. Five independent runs were executed for each setting.}
\label{fig:pythia_6_9b_ensemble_tpr}
\vspace{-0.1in}
\end{figure}

\begin{figure}[htb!]
\centering
\includegraphics[width=0.45\textwidth]{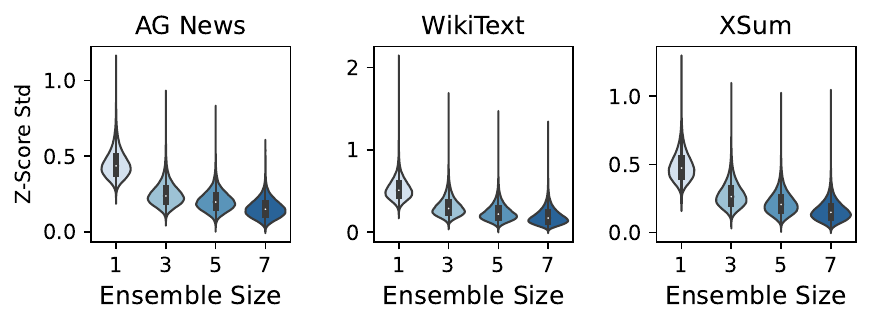}
\caption{Distribution of the standard deviation of z-scores computed from five independent runs of our method with varying ensemble sizes.}
\label{fig:pythia_6_9b_ensemble_z_score}
\vspace{-0.2in}
\end{figure}

Figure~\ref{fig:pythia_6_9b_ensemble_tpr} shows the results achieved by our method using varying ensemble configurations on three datasets, where the target model is Pythia-6.9b. 
We observe that performance improves when ensemble size increases in general. 
The variance in true positive rates from different runs of our method tends to decrease when the ensemble size increases.
The performance at 1\% FPR stabilizes for ensemble size 5 while there is some fluctuation at 0.1\% FPR. 
This can be explained by the fact the threshold corresponding to 0.1\% FPR is more sensitive to noise as the number of samples in consideration is less. 
Figure~\ref{fig:pythia_6_9b_ensemble_z_score} shows the distribution of standard deviation of z-scores computed from different runs of our method with varying ensemble sizes.
With an increased ensemble size, we observe variance in the computed z-scores from different runs reduces among the samples from the three datasets.
As a result, the noise in our prediction is reduced, which leads to better performance from ensembling. 

\subsection{Effect of Training Epoch and Model Size}

The preceding results all showed MIA performance against a single epoch of fine-tuning, since that is the harder setting to attack\footnote{one could argue that fractional epoch training is single epoch training on a reduced $\Dp$ dataset}. In this section, we study the how training epochs and size of target model affects privacy risk captured by different methods.

\begin{figure}[htb!]
\vspace{-0.15in}
\centering
\includegraphics[width=0.4\textwidth]{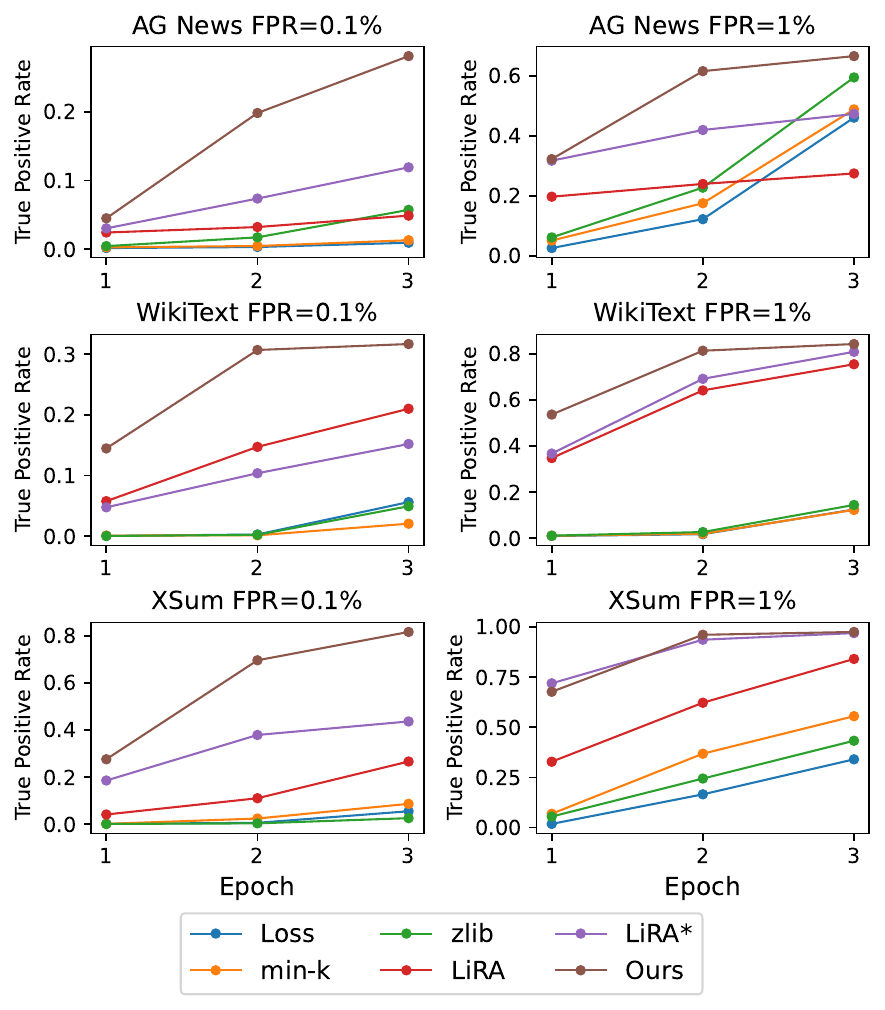}
\caption{True positive rates at 0.1\% and 1\% FPR on all datasets as a function of number of epochs of the target model (OPT-6.7b). MIA risk increases for all methods with additional fine-tuning epochs of the target model. }
\label{fig:opt_6_7b_epoch_tpr}
\vspace{-0.1in}
\end{figure}

Figure~\ref{fig:opt_6_7b_epoch_tpr} shows results when the target model is OPT-6.7b with varying epochs of fine-tuning.
All the methods achieve higher true positive rates when the target model is trained with more epochs, indicating an increase in the privacy risks associated. This finding is consistent with the findings in \cite{duan2024membership}.
While the performance of simple score-function-based methods are relatively poor on models trained for one epoch, they become more competitive when the number of epochs increases. 
Among the methods with per-sample calibration, our method consistently achieves better performance at 0.1\% FPR and comparable performance at 1\% FPR to variants of LiRA.

\begin{figure}[htb!]
\vspace{-0.1in}
\centering
\includegraphics[width=0.4\textwidth]{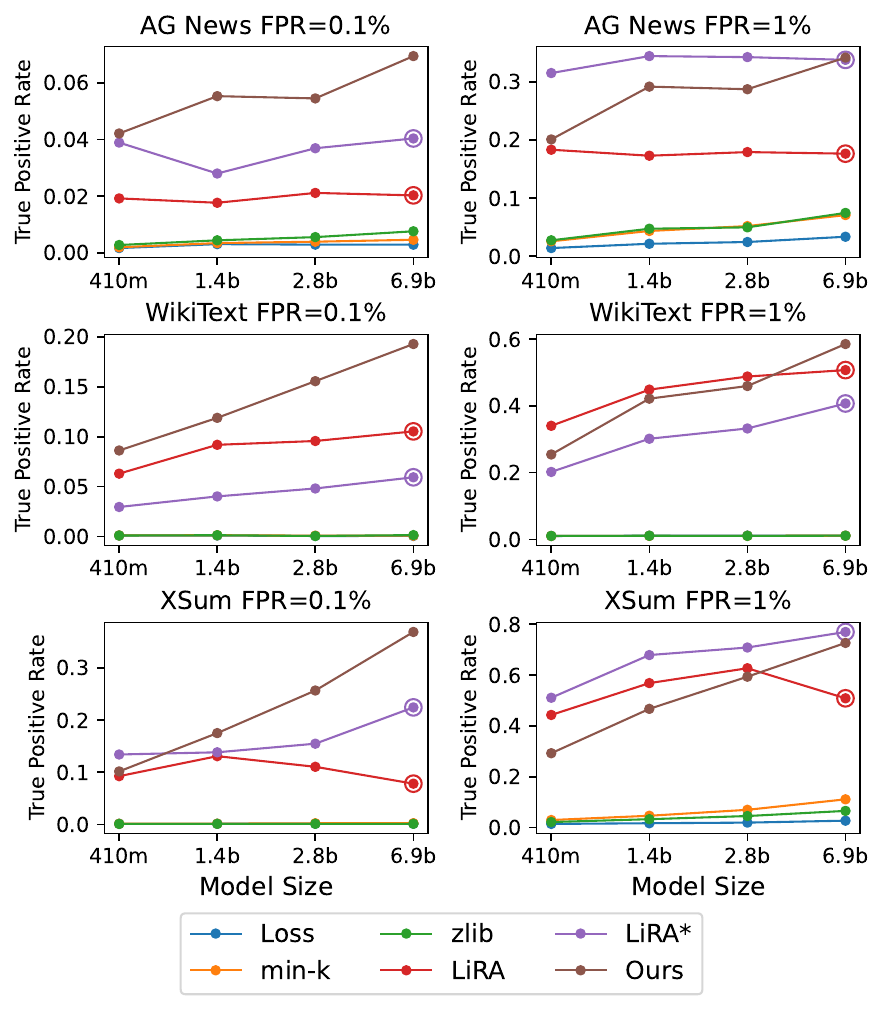}
\caption{True positive rates at 0.1\% and 1\% FPR on the three datasets where all target models Pythia models of different sizes. LiRA results obtained using shadow models of smaller sizes than target models are marked with empty circles.}
\label{fig:pythia_size_tpr}
\vspace{-0.15in}
\end{figure}

Figure~\ref{fig:pythia_size_tpr} shows the results when using Pythia models of different sizes as the target model.
For most methods, the true positive rates at both 0.1\% and 1\% FPR increases with target model size, which is in line with the findings on the impact of model size on memorization~\cite{carlini2021extracting}. 
LiRA exhibits more fluctuation in the performance, especially at the 0.1\% FPR. This is likely to due to the noise involved in the training of shadow models and mismatch in shadow model sizes for Pythia-6.9b target models.
On the other hand, our method is able to consistently capture the trend and obtain better true positive rates at 0.1\% FPR.

\section{Conclusion}
We have developed a membership inference attack methodology for large language models that is more computationally efficient than the prior state of the art by almost two orders of magnitude without sacrificing effectiveness --- in all experiments the accuracy of our attack either exceeds or is comparable to that of LiRA. These efficiency improvements are especially important when MIAs are intended to be used as a routine privacy auditing procedure for large deployed models, as shadow model attacks which are several times more expensive to run than the training of the target model itself become prohibitive.  
\section{Limitations}

As with all membership inference attacks, when using the attack we propose here as an auditing mechanism, it should be viewed as akin to a unit-test; susceptibility to the attack proposed here is a clear red flag, but lack of susceptibility does not provide any guarantee that the model is robust to other, yet-to-be-discovered privacy attacks. When provable guarantees are needed, techniques like differential privacy \cite{dwork2014algorithmic} should be employed.

\section{Broader Impact}

Advancements in membership inference attacks (MIAs) for large language models (LLMs) are important for improving privacy auditing and compliance. By improving the efficiency of MIAs, our work helps auditors more routinely evaluate deployed models for privacy properties (or lack thereof). By making privacy leakage more easily measurable, we hope our work encourages privacy to become a first-order design desiderata in large-scale machine learning. 

Improved MIAs of course also increase the risk of external attacks. Thus in the short run, work on privacy attacks (including ours) can increase the privacy risk of deployed models. Nevertheless, we believe that in the long run exposing privacy risk is an essential step to mitigating it.

\bibliography{main}

\clearpage
\appendix

\section{Comparison of Scoring Functions}
\label{sec:scoring_comparison}
In previous sections, we have compared our method with baselines leveraging different scoring functions, now we explore the performance of different scoring functions when per-sample based calibration is applied.
Table~\ref{tab:score_function} shows a comparison of different scoring functions and their counterparts where calibration is applied through LiRA and our method.
We observe that while min-k and zlib entropy typically improves over loss attack, their performance advantage is not necessarily maintained when calibrated using LiRA or our method.
This can potentially be attributed to the original design intent of these scoring functions, which were designed to not require (as much) calibration, and therefore benefit less from it. For instance, uncalibrated zlib entropy performs quite differently from the base loss attack, but their calibrated performances under LiRA are near identical. Calibration using our method has the best performance in most cases, especially at lower false positive rate.
Still, we noticed overall higher training losses on our regression models when using zlib entropy or min-k score as the scoring function, which may explain the worse performance with zlib entropy on XSum.

\begin{table}[h!]
\scriptsize	
\centering
\begin{tabular}{l@{\hspace{0.8em}}c@{\hspace{0.7em}}cc@{\hspace{0.7em}}cc@{\hspace{0.7em}}c}
\toprule
Dataset & \multicolumn{2}{c}{AG News} & \multicolumn{2}{c}{WikiText} & \multicolumn{2}{c}{XSum}\\
\cmidrule(l{0pt}r{4pt}){2-3} \cmidrule(l{4pt}r{4pt}){4-5} \cmidrule(l{4pt}r{4pt}){6-7}
FPR & 0.1\% & 1\% & 0.1\% & 1\% & 0.1\% & 1\%\\
\midrule
Loss & 0.29 & 3.35 & 0.04 & 1.09 & 0.15 & 2.70\\
min-k & 0.46 & 7.08 & 0.07 & 1.11 & 0.25 & 11.10\\
zlib & 0.76 & 7.46 & 0.12 & 1.12 & 0.13 & 6.55\\
\midrule
LiRA (loss/zlib) & 2.03 & 17.65 & 10.53 & 50.72 & 7.82 & 50.87\\
LiRA (min-k) & 1.43 & 16.71 & 9.22 & 44.33 & 8.67 & 47.50\\
\midrule
LiRA* (loss) & 4.04 & 33.78 & 5.92 & 40.72 & 22.45 & 76.99\\
LiRA* (min-k) & 1.81 & 34.73 & 6.99 & 38.44 & 24.68 & \textbf{78.70}\\
LiRA* (zlib) & 4.49 & \textbf{35.46} & 7.04 & 47.16 & 26.09 & 78.40\\
\midrule
Ours (loss) & 6.95 & 34.17 & \textbf{19.29} & \textbf{58.52} & 36.92 & 72.68\\
Ours (min-k) & \textbf{7.39} & 34.18 & 17.90 & 52.60 & \textbf{38.49} & 72.31\\
Ours (zlib) & 5.73 & 34.84 & 17.35 & 57.59 & 25.93 & 56.05\\
\bottomrule
\end{tabular}
\caption{\label{tab:score_function}True positive rates (\%) at 0.1\% and 1\% FPR on the three datasets where target model is Pythia-6.9b. Different scoring functions are used for LiRA and our method. LiRA results are obtained with 4 Pythia-2.8b shadow models.}
\vspace{-0.15in}
\end{table}

\section{Extended ROC Curve Results}
\label{sec:extended_roc_results}

\begin{figure}[thb!]
\centering
\includegraphics[width=0.475\textwidth]{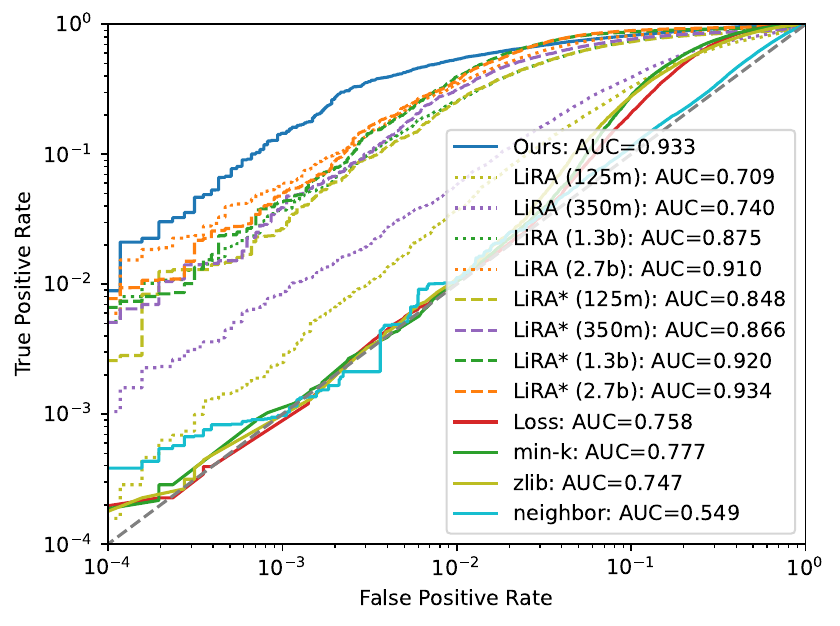}
\caption{Comparing true positive rates vs false positive rates of our method with LiRA variants and marginal baselines with different scoring functions on WikiText-103 where target model is OPT-6.7b. LiRA* represents LiRA with fixed variance estimate. 
Results for LiRA are obtained with 4 shadow models from OPT family with varying sizes. 
Results for our method are obtained with ensemble of 5 quantile regression models finetuned from opt-125m.
}
\label{fig:opt_6_7_b_wikitext_roc}
\vspace{-0.15in}
\end{figure}

\begin{figure}[thb!]
\centering
\includegraphics[width=0.475\textwidth]{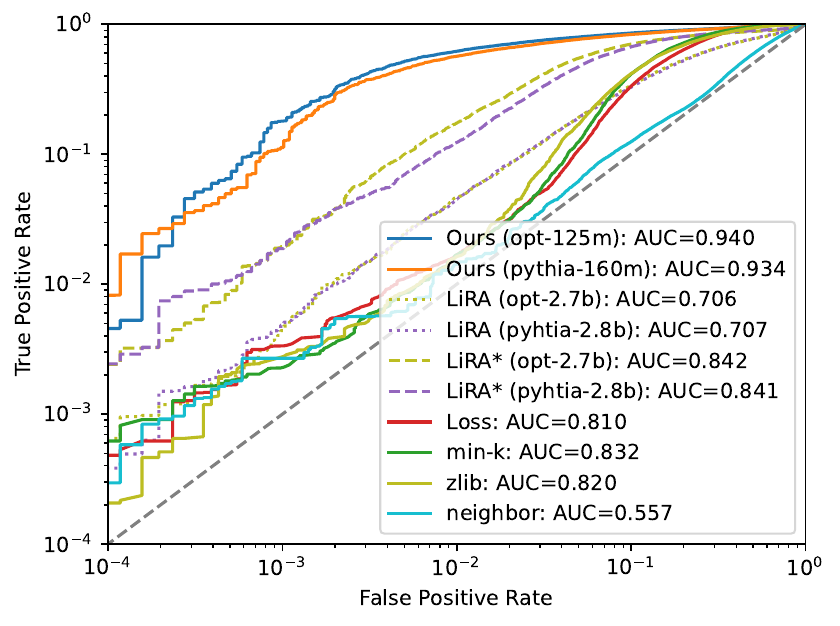}
\caption{Comparing true positive rates vs false positive rates of our method with LiRA variants and marginal baselines with different scoring functions on WikiText-103 where target model is Llama-7b. LiRA* represents LiRA with fixed variance estimate. 
Results for LiRA are obtained with 4 shadow models from OPT-2.7b and Pythia-2.8b. 
Results for our method are obtained with ensemble of 5 quantile regression models finetuned from OPT-125m and Pythia-160m.
}
\label{fig:llama_7b_wikitext_roc}
\vspace{-0.15in}
\end{figure}

\begin{figure}[tb!]
\centering
\includegraphics[width=0.475\textwidth]{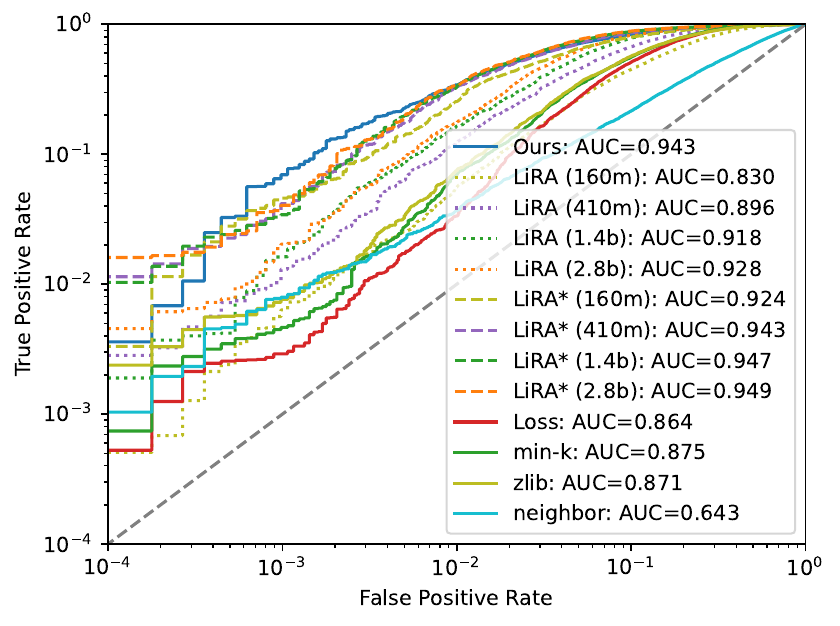}
\caption{Comparing true positive rates vs false positive rates of our method with LiRA variants and marginal baselines with different scoring functions on AG News where target model is Pythia-6.9b. LiRA* represents LiRA with fixed variance estimate. Results for LiRA are obtained with 4 shadow models from Pythia family with varying sizes. Results for our method are obtained with ensemble of 5 quantile regression models finetuned from Pythia-160m.}
\label{fig:pythia_6_9_b_ag_news_roc}
\vspace{-0.15in}
\end{figure}

\begin{figure}[tb!]
\centering
\includegraphics[width=0.475\textwidth]{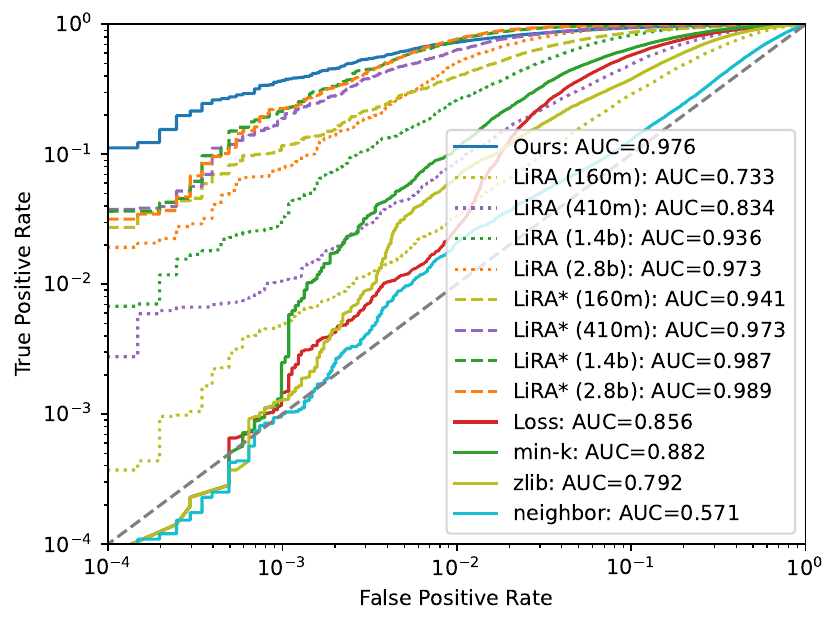}
\caption{Comparing true positive rates vs false positive rates of our method with LiRA variants and marginal baselines with different scoring functions on XSum where target model is Pythia-6.9b. LiRA* represents LiRA with fixed variance estimate. Results for LiRA are obtained with 4 shadow models from Pythia family with varying sizes. Results for our method are obtained with ensemble of 5 quantile regression models finetuned from Pythia-160m.}
\label{fig:pythia_6_9_b_xsum_roc}
\vspace{-0.15in}
\end{figure}

Here we show extended ROC curve results of our experiments. 
Figure~\ref{fig:opt_6_7_b_wikitext_roc} and Figure~\ref{fig:llama_7b_wikitext_roc} show ROC curves on WikiText where target models are OPT-6.7b and Llama-7b.
Figure~\ref{fig:pythia_6_9_b_ag_news_roc} and Figure~\ref{fig:pythia_6_9_b_xsum_roc} show ROC curves on AG News and XSum where target models are Pythia-6.9b.

\section{Varying Sizes of Regression Models}
\label{sec:extended_size_results}
Here we study how the size of quantile regression models affects the performance of our method.
Table~\ref{tab:qr_size} shows a performance comparison of our method using Pythia models of varying sizes for regression when the target model is Pythia-6.9b.
We observe an improvement of true positive rates at 0.1\% and 1\% FPR as regression model size increases.
In our experiments, we observed lower regression loss when training with larger models, which may explain the improvement in performance.

\begin{table}[thb!]
\scriptsize	
\centering
\begin{tabular}{l@{\hspace{0.8em}}c@{\hspace{0.7em}}cc@{\hspace{0.7em}}cc@{\hspace{0.7em}}c}
\toprule
Dataset & \multicolumn{2}{c}{AG News} & \multicolumn{2}{c}{WikiText} & \multicolumn{2}{c}{XSum}\\
\cmidrule(l{0pt}r{4pt}){2-3} \cmidrule(l{4pt}r{4pt}){4-5} \cmidrule(l{4pt}r{4pt}){6-7}
FPR & 0.1\% & 1\% & 0.1\% & 1\% & 0.1\% & 1\%\\
\midrule
Loss & 0.29 & 3.35 & 0.04 & 1.09 & 0.15 & 2.70\\
\midrule
LiRA (2.8b) & 2.03 & 17.65 & 10.53 & 50.72 & 7.82 & 50.87\\
LiRA* (2.8b) & 4.04 & 33.78 & 5.92 & 40.72 & 22.45 & 76.99\\
\midrule
Ours (70m) & 5.74 & 30.54 & 11.48 & 49.24 & 30.30 & 65.12 \\
Ours (160m) & 6.95 & 34.17 & 19.29 & 58.52 & 36.92 & 72.68\\
Ours (410m) & \textbf{7.84} & \textbf{40.59} & \textbf{22.00} & \textbf{63.89} & \textbf{40.38} & \textbf{78.24} \\
\bottomrule
\end{tabular}
\caption{\label{tab:qr_size}True positive rates (\%) at 0.1\% and 1\% false positive rates on the three datasets; the target model is Pythia-6.9b. LiRA results are obtained with 4 shadow models. LiRA* represents LiRA with fixed variance. Results for our method are obtained using ensembles of 5 models from finetuning Pythia models of different sizes.}
\end{table}

\section{Varying Sizes of Regression Models}
\label{sec:extended_size_results}
Here we study how the size of quantile regression models affects the performance of our method.
Table~\ref{tab:qr_size} shows a performance comparison of our method using Pythia models of varying sizes for regression when the target model is Pythia-6.9b.
We observe an improvement of true positive rates at 0.1\% and 1\% FPR as regression model size increases.
In our experiments, we observed lower regression loss when training with larger models, which may explain the improvement in performance.

\section{Extended Comparison with LiRA Using Increased Shadow Models}
Here we present extended comparison with LiRA variants using increased number of shadow models. 
Table~\ref{tab:cross_family_increased_shadow} shows a comparison of our method with LiRA on WikiText where the model family varies among target model and attacker models.
We observe improved performance of LiRA with variable variance with the increased shadow models. 
Our method is able to achieve competitive results with a fraction of the time required for preparing the attacker models and consistently outperforms LiRA when the target model is Llama-7b.

\begin{table}[h!]
\scriptsize	
\centering
\begin{tabular}{@{\hspace{0.2em}}l@{\hspace{0.8em}}c@{\hspace{0.8em}}c@{\hspace{1.0em}}c@{\hspace{0.8em}}c@{\hspace{1.0em}}c@{\hspace{0.8em}}c@{\hspace{0.9em}}c@{\hspace{0.2em}}}
\toprule
Model & \multicolumn{2}{@{\hspace{-0.8em}}c@{}}{Pythia-6.9b} & \multicolumn{2}{@{\hspace{-0.6em}}c@{}}{OPT-6.7b} & \multicolumn{2}{@{\hspace{-0.6em}}c@{}}{Llama-7b}& Time \\
\cmidrule(l{0pt}r{6pt}){2-3} \cmidrule(l{0pt}r{6pt}){4-5} \cmidrule(l{0pt}r{6pt}){6-7}
FPR & 0.1\% & 1\% & 0.1\% & 1\% & 0.1\% & 1\% & (hrs) \\
\midrule
Loss Attack & 0.04 & 1.09 & 0.04 & 0.93 & 0.33 & 1.68 & - \\
\midrule
LiRA (P-2.8b n=8) & \textbf{28.93} & \textbf{72.32} & \underline{15.25} & 51.78 & 1.83 & 11.46 & 110.0 \\
LiRA (O-2.7b n=8) & 13.38 & 55.31 & \textbf{15.46} & \textbf{61.43} & 1.18 & 9.51 & 116.8 \\
\midrule
LiRA* (P-2.8b n=8) & 5.77 & 42.08 & 4.13 & 21.66 & 2.14 & 12.57 & 110.0 \\
LiRA* (O-2.7b n=8) & 5.56 & 46.27 & 4.27 & 37.42 & 2.01 & 17.42 & 116.8 \\
\midrule
Ours (P-160m) & 19.29 & 58.52 & 13.03 & 45.98 & 11.28 & 56.74 & 3.4 \\
Ours (O-125m) & 18.03 & 61.19 & 14.48 & 53.69 & \underline{17.96} & \underline{61.87} & 3.4 \\
Ours (P-410m) & 22.00 & 63.89 & 13.92 & 54.31 & \textbf{18.95} & 61.84 & 11.4 \\
Ours (O-350m) & \underline{22.55} & \underline{66.74} & 14.84 & \underline{56.95} & 17.00 & \textbf{63.74} & 11.4 \\

\bottomrule
\end{tabular}
\caption{\label{tab:cross_family_increased_shadow} True positive rates (\%) at 0.1\% and 1\% FPR with different target models on the WikiText dataset along with the total time to prepare the shadow models or regression models. P-2.8b, -160m, -410m and O-2.7b, -125m, -350m correspond to Pythia-2.8b, -160m, -410m and OPT-2.7b, -125m, 350m, respectively. LiRA results are obtained with 8 shadow models from the Pythia and OPT families. LiRA* represents LiRA with fixed variance. Results for our method are obtained using ensembles of 5 regression models.}
\vspace{-0.15in}
\end{table}



\end{document}